\documentclass[letterpaper, 10 pt, conference]{ieeeconf}  

\IEEEoverridecommandlockouts                              

\let\labelindent\relax
\usepackage{cite}
\usepackage{graphicx}
\usepackage{subfigure}
\usepackage{dblfloatfix}
\usepackage{amsmath}
\usepackage{booktabs}
\usepackage{multirow}
\usepackage{enumitem}
\usepackage{amsfonts}
\usepackage{CJK}
\usepackage{verbatim}
\usepackage{hyperref}

\title{\LARGE \bf
D3-ARM: High-Dynamic, Dexterous and Fully Decoupled 

Cable-driven  Robotic Arm
}

\author{Hong Luo, Jianle Xu, Shoujie Li, Huayue Liang, Yanbo Chen, Chongkun Xia and Xueqian Wang 
\thanks{This work was supported by the the National Key R\&D Program of China (2022YFB4701400/4701402), National Natural Science Foundation of China (No.62103225, U21B6002, 62203260, 92248304), Natural Science Foundation of Shenzhen (No. JCYJ20230807111604008), Natural Science Foundation of Guangdong Province (No.2024A1515010003) and Guangdong Basic and Applied Basic Research Foundation (2023A1515011773).}
\thanks{Hong Luo, Jianle Xu, Shoujie Li, Huayue Liang, Yanbo Chen and Xueqian Wang are with the Center for Artificial Intelligence and Robotics, Shenzhen International Graduate School, Tsinghua University, Shenzhen 518055, China.}
\thanks{Chongkun Xia is with School of Advanced Manufacturing, Sun Yat-sen University, shenzhen 518107, China.}
\thanks{Corresponding author: Chongkun Xia (xiachk5@mail.sysu.edu.cn), Xueqian Wang (wang.xq@sz.tsinghua.edu.cn)}
}

\begin{document}

\maketitle
\thispagestyle{empty}
\pagestyle{empty}

\begin{abstract}


Cable transmission enables motors of robotic arm to operate lightweight and low-inertia joints remotely in various environments, but it also creates issues with motion coupling and cable routing that can reduce arm's control precision and performance.
In this paper, we present a novel motion decoupling mechanism with low-friction to align the cables and efficiently transmit the motor's power. By arranging these mechanisms at the joints, we fabricate a fully decoupled and lightweight cable-driven robotic arm called D3-Arm with all the electrical components be placed at the base. 
Its 776 mm length moving part boasts six degrees of freedom (DOF) and only 1.6 kg weights. To address the issue of cable slack, a cable-pretension mechanism is integrated to enhance the stability of long-distance cable transmission. Through a series of comprehensive tests, D3-Arm demonstrated 1.29 mm average positioning error and 2.0 kg payload capacity, proving the practicality of the proposed decoupling mechanisms in cable-driven robotic arm. 


\end{abstract}

\section{INTRODUCTION}

\begin{figure}[t]
    \centering
    \includegraphics[width=1\linewidth]{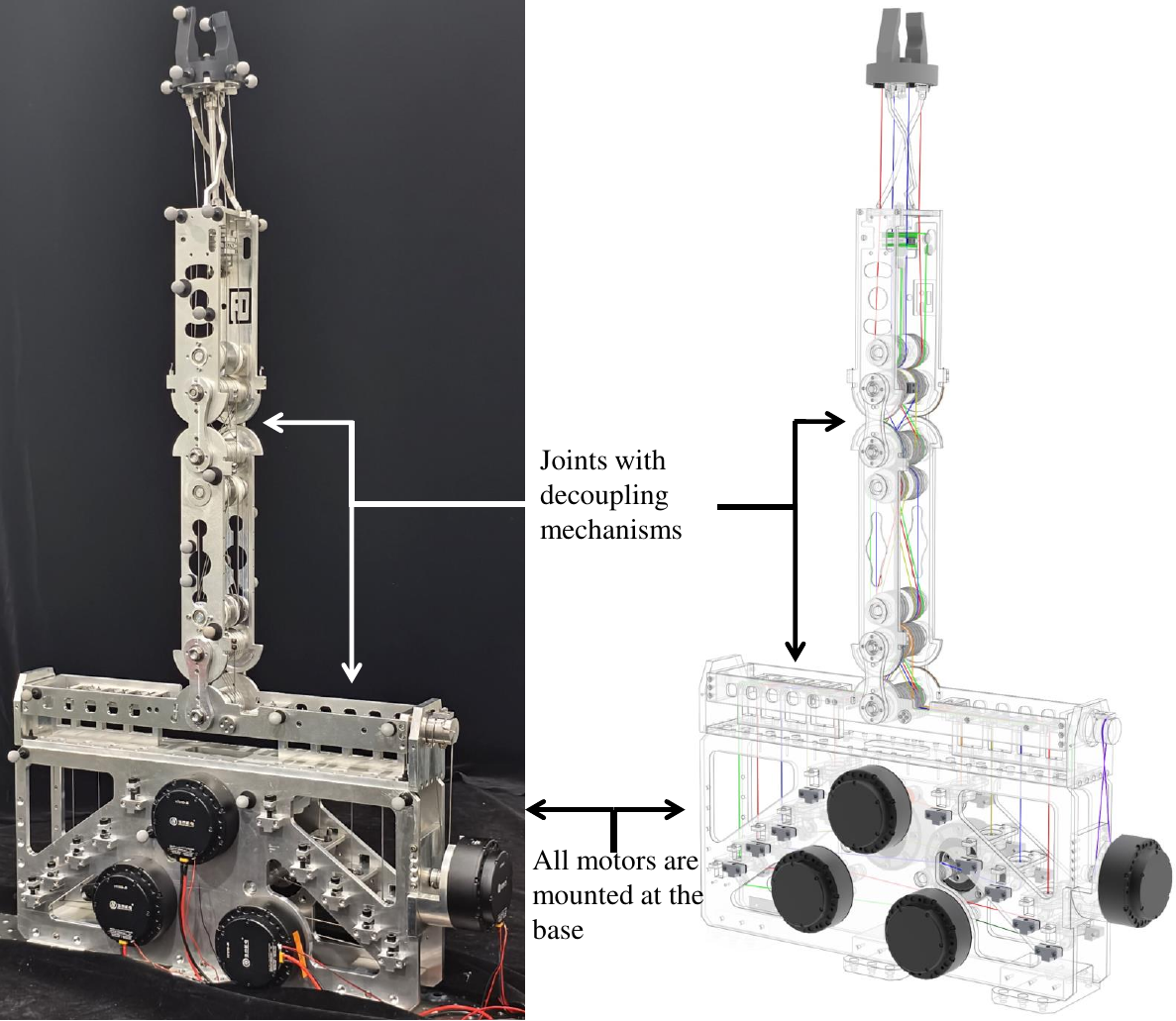}
    \caption{D3-Arm: a fully decoupled cable-driven robotic arm with all its electrical components located at the base.}
    \label{fig:enter-label}
    \vspace{-0.5cm}
\end{figure}

With the features of backlash-free actuation and remote transmission, cable-driven joints are widely used in the lightweight design of robotic arms by placing motors at the proximal part\cite{lightweight_cb_manipulator,ozawa2013analysis,6386236}. However, for certain environments such as high radiation and underwater tasks, centralized protection of components like motors is necessary to enhance the durability like radiation resistance of the robotic arms\cite{jiang2013mechanism}. This necessitates that all motors of the cable-driven arm be housed within the base to ensure isolation from the operating environment\cite{Dragon,6907730}. In this paper, we design a cable-driven robotic arm suitable for such environments, as illustrated in Fig. \ref{fig:enter-label}.


To concentrate all the motors of the cable-driven robotic arm at the base and effectively drive the joints, the primary challenge is to design an efficient cable transmission routing. One of the most straightforward methods is to pass the driving cables directly through holes in the joint components and arrange these modular joints in a series to form a cable-driven continuum arm\cite{continuum1,continuum2,continuum3,continuum4} or use tubes to constrain the transmission path of the cables\cite{Self-calibration,self-calibration_2,bowden}. Despite its simplicity, this transmission method can introduce significant friction influence, which in turn reduces the control precision and efficiency of the cable-driven arm\cite{Parallel_Continuum}. To avoid this friction issues, grooved pulleys are widely used along the cable transmission path\cite{wam,constant_tendon_length}. The Twist Snake \cite{snake} designed a joint composition where multiple pulleys shared one joint axis to send the cables from motors at the base to driven joints. SAQIEL\cite{SAQIEL} introduced a passive 3D wire alignment mechanism composed by grooved pulleys to achieve a supple drive from actuators onto the root link by minimal friction. This transmission method can effectively solve the energy loss issue along the long transmission path of cable-driven arms.

\begin{figure*}[t] 
    \centering
    \includegraphics[width=1\linewidth]{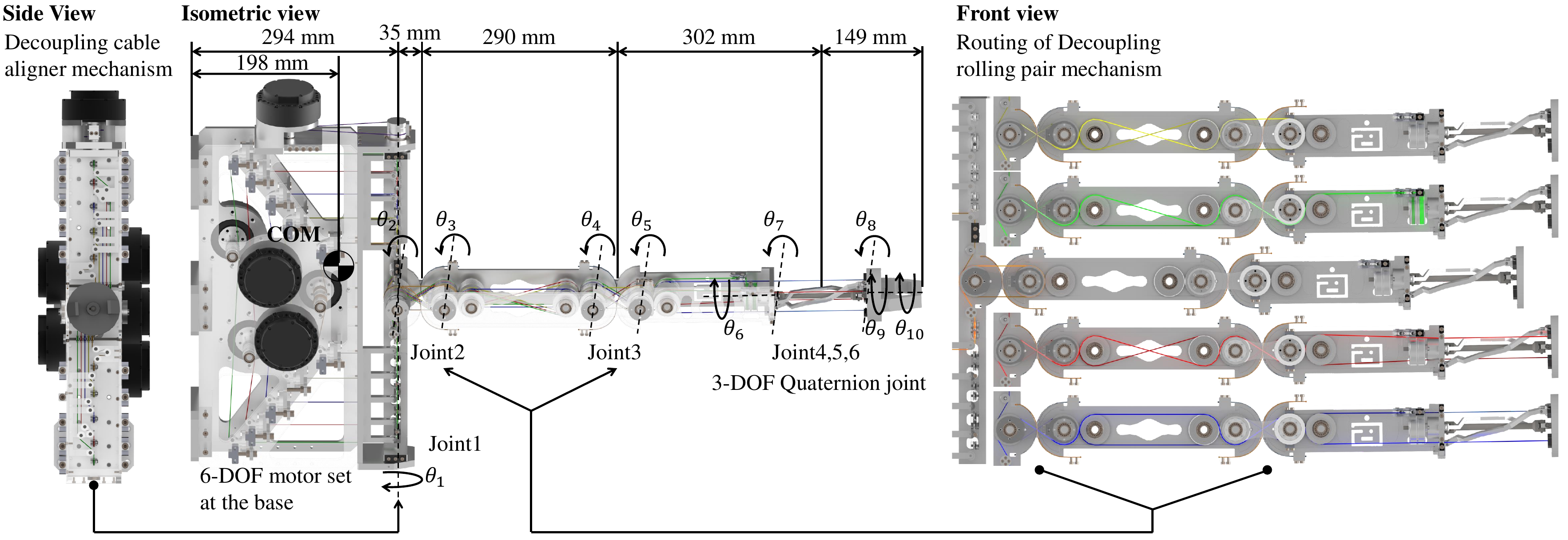}
    \caption{Proportion and joint composition of D3-Arm. Pairs of lines in the same color indicate a group of cables that drive the same joint. The side view shows the cable routing of the decoupling cable aligner mechanism arranged in Joint1. The front view shows the cable routing of decoupling rolling pair mechanism arranged in Joint2 and Joint3.}
    \label{fig:Proportion}
    \vspace{-0.5cm}
\end{figure*}
However, the abovementioned cable transmission methods inevitably bring coupling problems where the movement of the certain link can affect the length and tension of the cables driving other links. This deficiency greatly limits the control precision and durability of the robotic arm\cite{motion_decoupled}. To solve this coupling problem, many research efforts have focused on compensating the length variations in cable coupling by implementing decoupling mechanisms\cite{jiang2018design,choi2024design}. The follower decoupling mechanism that synchronously drags the cables with joint movement ensures the length of the cable driving other joints remain constant\cite{constant_length}. The alignment decoupling mechanism, which keeps the cable consistently aligned with the joint axis, is also an effective method to prevent coupling effects on the cable\cite{RoboRay_hand}. Based on these strategies, the LIMS arm used a rolling joint as its elbow, successfully decoupling the elbow motion and the movement of the wrist tendon\cite{kim,kim2018development}. The CDSSR achieved full-joint decoupling through the cable length compensator in each articulation\cite{xu2024design}. However, this compensation-based decoupling method adversely affects cable tension stability, thus limiting joint dynamic performance. While the CDSSR demonstrates the feasibility of full-arm decoupling, achieving high-speed and fully decoupled motion still requires further improvements in transmission system design.

In this paper, we aim to develop a fully decoupled cable-driven robotic arm with all its motors mounted at the base. To achieve this goal, we propose lightweight cable-driven joints equipped with decoupling mechanisms, utilizing a cable-pulley configuration to facilitate low-friction power transmission, and assemble these joints in series to form a 6 DOF cable-driven robotic arm. This cable-driven robotic arm can not only facilitate the isolation and protection of components such as motors to adapt to various environments, its fully decoupled system also enhance the control precision and sufficient dynamic performance. The main contributions of this work are as follows:

\begin{enumerate}
    \item A fully decoupled and low-friction transmission mechanism is proposed to enable high-efficiency and high-precision remote transmission via cables.
    \item A cable-driven arm is designed based on the decoupling mechanisms, with all the electrical components such as motors mounted at the base, ensuring the operational safety in interactive or extreme environments.
    \item The effectiveness of the decoupling mechanisms in enhancing the control precision of the cable-driven arm is validated through end-effector position repeatability experiment.
\end{enumerate}

\section{Mechanism Design of D3-Arm}

\subsection{Design Requirement} 
In this chapter, we provide a detailed description for the design of the 6-DOF cable-driven robotic arm named D3-Arm(High-\textbf{D}ynamic,\textbf{D}exterous and fully \textbf{D}ecoupled).

From the aforementioned research, it is clear that to enhance the durability of the cable-driven arm in extreme environments, the isolation for the electrical components like motors from the operating environment is essential. 
To achieve this configuration, it is necessary to design a high-efficiency and fully decoupled cable transmission path to enhance remote control accuracy, which implies the integration of low-friction and lightweight decoupling mechanisms within the joints. In addition, the decoupling mechanism within each joint must be able to simultaneously decouple multiple driving cables to enhance the flexibility of the robotic arm. Finally, to fully leverage the advantage of the remote driving feature of cables, the mass and inertia of the robotic arm's moving parts should be sufficiently low while ensuring a certain load capacity to enhance its dynamic motion capabilities and collision safety. From this, the following design objectives for the D3-Arm can be summarized:

\begin{enumerate}
    \item \emph{Safety in multiple environments}: All motors and other electrical components are centrally arranged at the base, ensuring the durability and applicability in human-robot interactive or extreme environments.
    \item \emph{Accuracy in aforementioned configuration}: Implementing decoupling mechanisms in the cable routing to make all the joint motions be completely independent.
    \item \emph{Utilization of the features in cable-driven system}: Minimizing the friction in the cable transmission system and the inertia of moving parts to maximize the advantages of cable-driven systems.

\end{enumerate}

\subsection{Joint Composition}
Fig.\ref{fig:Proportion} shows the proportion and the joint composition of D3-Arm. The six driving motors are positioned at the base and remotely actuate the corresponding six joints through six pairs of driving cables. Among these, Joint1, Joint2, and Joint3 are rotational or rolling joints. To ensure that the robotic arm has a sufficiently large working space, the motion axis of Joint1 is arranged to be perpendicular to that of Joint2. In contrast, the remaining terminal joints employs a 3-DOF quaternion joint due to considerations of size and integration\cite{quternion}. This parallel joint not only allows for independent motions but also provides ample movement space, making it highly suitable as a end-effector for the fully decoupled cable-driven system.

In the transmission path, only the driving cable of Joint1 is directly connected to the motor, while the remaining driving cables must continuously pass through at least one joint before they can connect to the motors. In this configuration, Joint1, Joint2, and Joint3 require the design of decoupling mechanism to prevent the influence of joint motions on the driving cables. Besides, the decoupling mechanisms of those three joints should allow cable routing of at least six actuation cables, making the modular design of the mechanism extremely important. To this end, we have designed and implemented two types of decoupling mechanisms, referred to 1-DOF decoupling cable aligner mechanism and 1-DOF decoupling rolling pair mechanism.

\begin{figure}[t]
    \centering
    \includegraphics[width=1\linewidth]{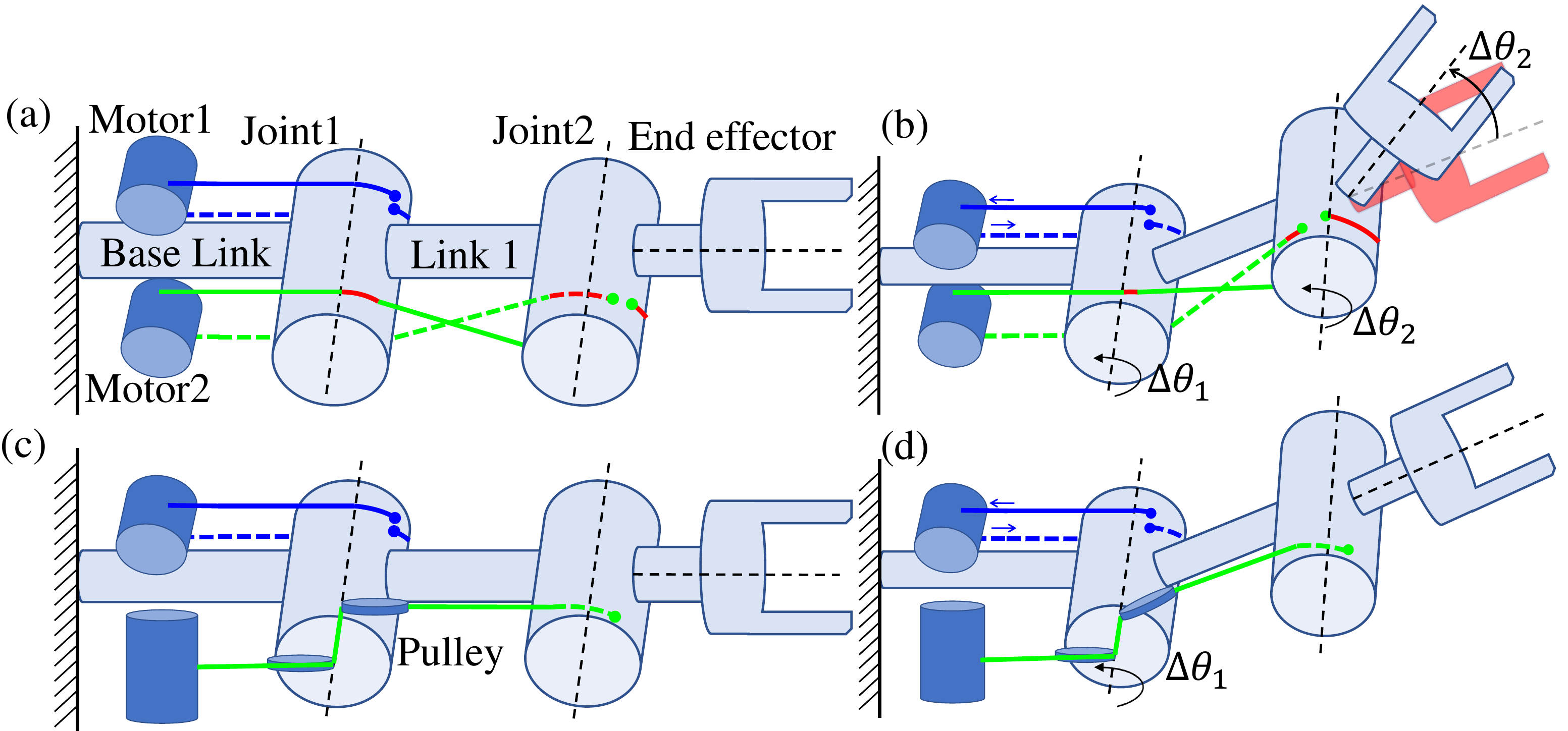}
    \caption{Schematics of the motion coupling problem and decoupling way during a pitching motion. (a) Normal routing of cables in two adjoining joints. (b) Motion coupling in normal routing. (c) Decoupled routing of cables in two adjoining joints. (d) Motion situation after decoupling.}
    \label{fig:Schematics}
    \vspace{-0.5cm}
\end{figure}

\subsection{Decoupling Cable Aligner Mechanism}
The coupling problems caused by remote driving with the cables during pitch motion are shown in the Fig. \ref{fig:Schematics}(a). The driving cables of Joint2 need to pass through Joint1 before connecting to Motor2, causing its winding section on Joint1 to become a coupled part (highlighted in red in Fig. \ref{fig:Schematics}(a)). When Motor1 drives Joint1 to move by an angle $\Delta \theta_1$, the length of the coupled portion of the driving cable for Joint2 change accordingly, causing Joint2 to also move by an angle $\Delta \theta_2$, even though Motor2 is not actively driving it. Finally, as shown in Fig. \ref{fig:Schematics}(b), this coupled angle causes the end effector to deviate from its originally intended position (the red position in Fig. \ref{fig:Schematics}(b)).

To solve this problem, an effective and easily extendable decoupling method is to align the driving cable with the motion axis of Joint1 using a pulley system (as shown in Fig. \ref{fig:Schematics}(c)). This system requires at least two grooved pulleys, one of which is rigidly connected to the base to tension and pull the cable to the motion axis of Joint1, while the other is rigidly connected to Link1, ensuring the cable remains aligned with the axis at any angle of movement as shown in Fig. \ref{fig:Schematics}(d). Thus, the motion of Joint1 does not affect the length of the driving cable for Joint2, achieving decoupled motion.

\begin{figure}[t]
    \centering
    \includegraphics[width=1\linewidth]{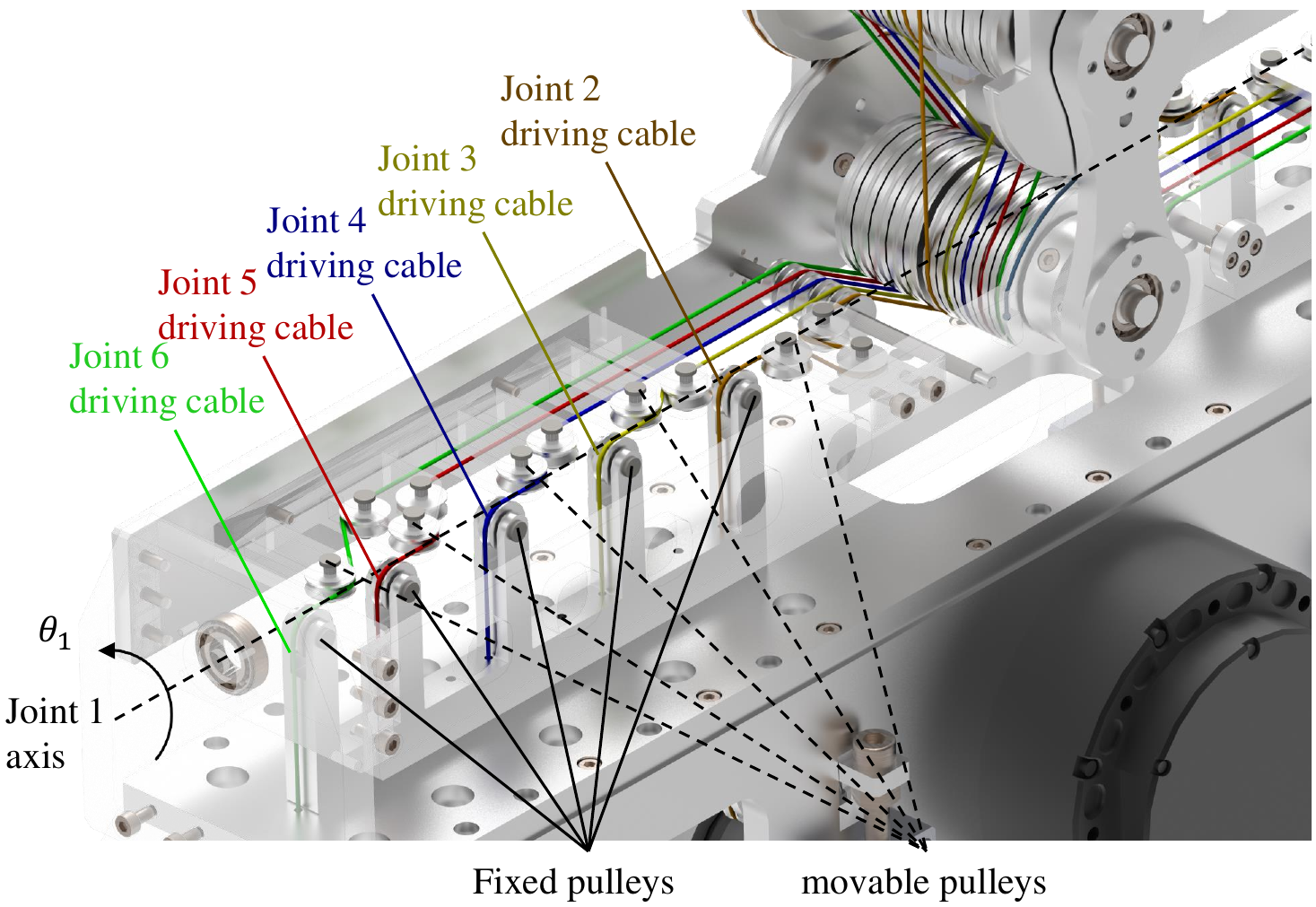}
    \caption{Detailed design of decoupling cable aligner Mechanism. The mechanism consists of fixed pulleys and movable pulleys, ensuring that the driving cables remain aligned with the motion axis of Joint1.}
    \label{fig:design of aligner}
    \vspace{-0.5cm}
\end{figure}

Based on the aforementioned principle, we design a decoupling cable aligner mechanism and applied it into Joint1 of the D3-Arm to decouple the 10 driving cables that pass through it, as shown in Fig. \ref{fig:design of aligner}. The mechanism is composed of fixed pulleys and movable pulleys, where the fixed pulleys are rigidly connected to the base, and movable pulleys are rigidly connected to the Joint1. They align the driving cables for five joints from the motors to the motion axis of Joint1. Considering the interference between this driving cables, it is necessary to arrange these pulleys sequentially and route the cables radially. The top view of the cable routing within the decoupled device is illustrated as shown in side view of Fig. \ref{fig:Proportion}, where the driving cables for the same joint are arranged symmetrically about the center to facilitate the subsequent configuration of the cables for other joints. With the correct cable routing and pulley configuration, the driving cables can always follow the motion axis of Joint1 during its rotational movement.

\subsection{Decoupling Rolling Pair Mechanism}
To address the issue of cable coupling in multi-joint systems, it is also essential to minimize the number of transmission components that the cables pass through, thereby reducing the risk of cable slackening and detachment\cite{snake}. Therefore, to accommodate the cable routings in Joint1, the decoupling mechanism in Joint2 and Joint3 need to decouple the driving cables distributed along their axial directions. 

\begin{figure}[t]
    \centering
    \includegraphics[width=1\linewidth]{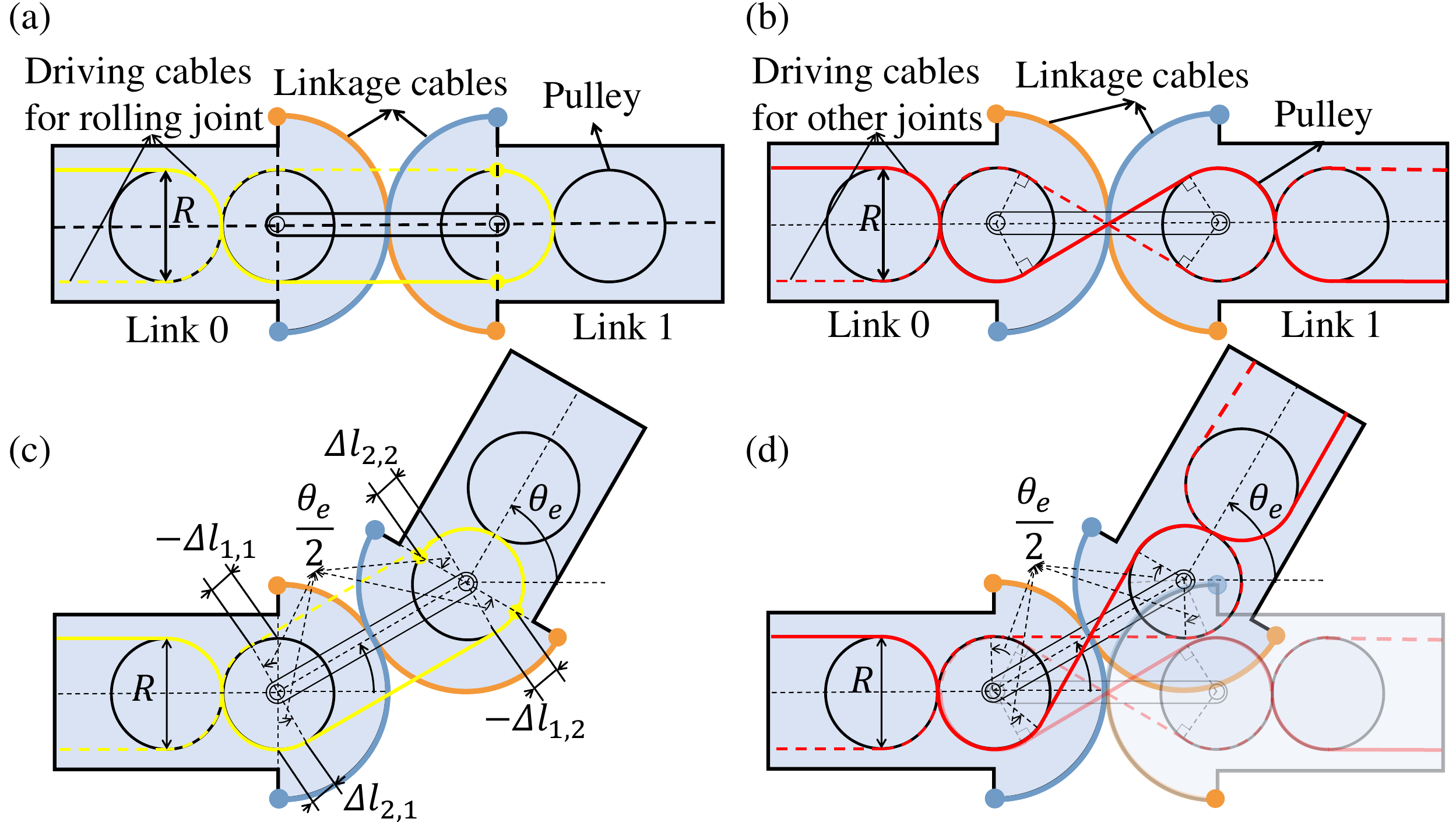}
    \caption{Detailed design of decoupled rolling pair Mechanism. (a) cable routings in joint motion. (b) cable routings in decoupling way. (c) Relationship between the cable motion and joint angle. (d) Cable displacement during joint movement.}
    \label{fig:rolling}
    \vspace{-0.5cm}
\end{figure}

In addition to the decoupling method that aligns the cables with the joint axes, another approach involves using rolling constraints to convert 1-DOF rotational joints into rolling joints. Under rolling contact, the motion of the joint does not affect the length of the cable passing through it. The cable routings under the rolling joint are shown in Fig. \ref{fig:rolling}(a) and Fig. \ref{fig:rolling}(b). Link0 and Link1 are articulated through an intermediate link and are subjected to rolling constraints by two linkage cables. The relationship between the cable motion and joint angle is shown in Fig. \ref{fig:rolling}(c), where the the solid and dashed yellow line represent the driving cables for the rolling joint. Based on the rolling constraint, the intermediate link and Link1 can move through the same angle $\theta_e/2$. Finally, the angle of Link1 relative to Link0 is related to the length of the driving cables $\Delta l_1$ by the following relationship:
\begin{equation}
\Delta l_1 = \Delta l_{1,1} + \Delta l_{1,2} = \frac{R\theta_e}{2}
\end{equation}
Similarly, the length change of another driving cable $l_2$ can also be determined as
\begin{equation}
\Delta l_2 = \Delta l_1 = \frac{R\theta_e}{2}
\end{equation}
When the rolling joint is in motion, the length changes of the driving cables for the remaining joints are illustrated in Fig. \ref{fig:rolling}(d). Since these cables can pass through the rolling contact point at any angle of joint movement, the length changes of the cable in the two pulleys always offset each other. Therefore, the length of the cables passing through this joint remains constant.

Similar to the advantages of the decoupling cable aligner mechanism, this decoupling mechanism enables multi-cable decoupling through axial expansion. We arrange this decoupling mechanism within Joint2 and Joint3 to decouple the 6 driving cables of the 3-DOF quaternion joint. The routing of the cables for each joint on the cable-driven arm is illustrated in the Front view of Fig. \ref{fig:Proportion}.

\subsection{Cable-pretension Module Mechanism}

The proposed decoupling mechanisms are all composed of pulleys to reduce frictional resistance during cable movement; however, this also increases the risk of cable being slack and coming off the pulleys\cite{8593679,9349131}. Additionally, as the motor drives the roller, the cable's axial position shifts, causing it to move out of the groove of pulleys.

\begin{figure}[t]
    \centering
    \includegraphics[width=1\linewidth]{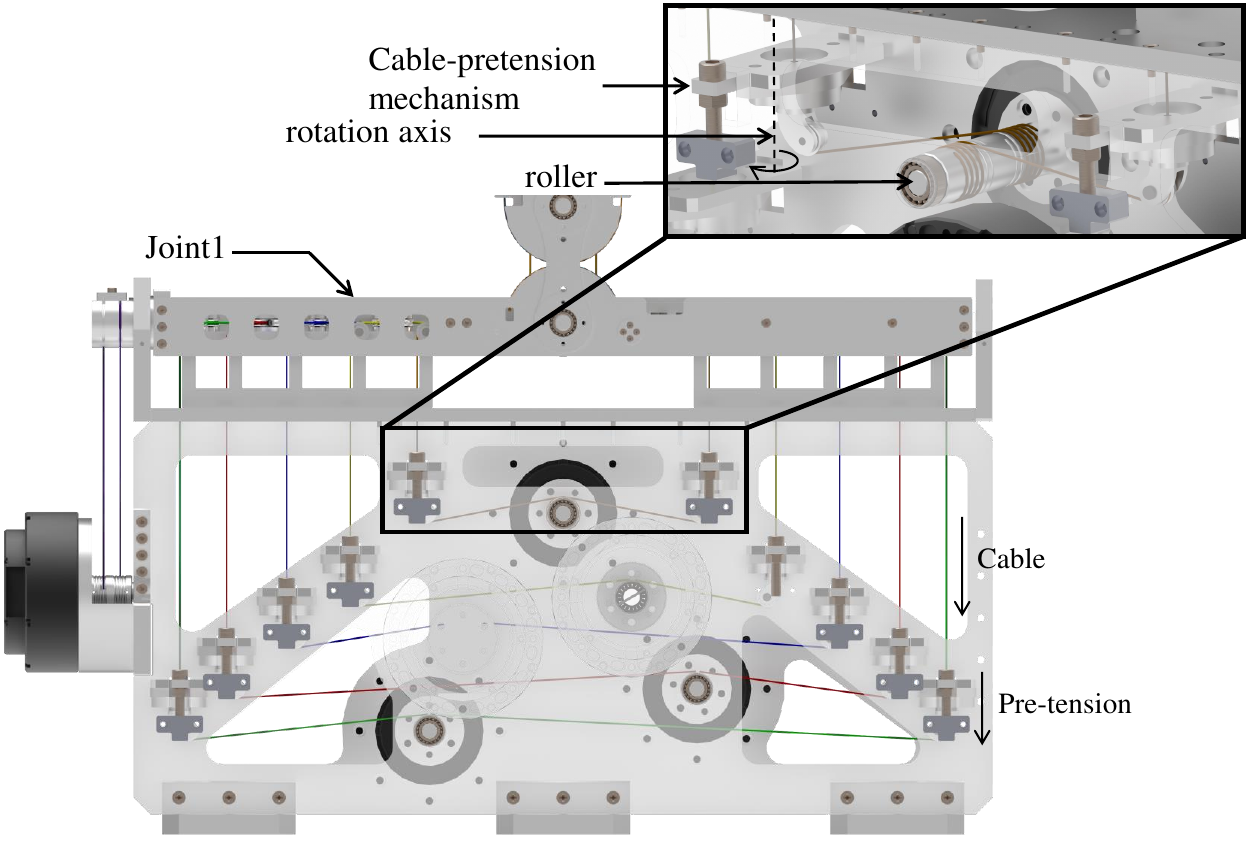}
    \caption{Detailed design of cable-pretension mechanism in the drive box. It can rotate around its rotation axis to draw the cable wrapped around the roller towards Joint1 and apply pre-tension force in accordance with the threaded direction.}
    \label{fig:cable-pretension}
    \vspace{-0.5cm}
\end{figure}

To address these issues, we design a cable-pretension mechanism within the drive box, as shown in Fig. \ref{fig:cable-pretension}. This mechanism provides pre-tension force through bolts and uses groove bearings that can rotate passively to guide the driving cables from the roller to the fixed pulleys of the decoupling mechanism in Joint1. Each driving cable is equipped with this cable-pretension mechanism and applied the same pre-tension force by utilizing an external force sensor.

\subsection{Overall Configuration}
Based on the above design, the primary specification of D3-Arm are presented in Table. \ref{tab:Specifications}. D3-Arm's length is 1.07 m with a approximately 1.6 kg weight moving part. To minimize the overall volume of the arm, a 1 mm steel wire rope is chosen as the driving cable. Based on the design principles, it can be inferred that selecting a steel cable of greater diameter would effectively enhance the stiffness and load-bearing capacity of the entire arm.

\begin{table}[ht]
    \centering
    \caption{Specifications of D3-Arm}
    \begin{tabular}{ccc} 
        \toprule 
        \multicolumn{2}{c}{Items} & Value \\ 
        \midrule 
        \multirow{2}{*}{Mass} & Overall arm & 16.5 kg\\ 
                              & Moving part & 1.6 kg \\ 
        \midrule 
        \multirow{6}{*}{Range of motion} & Joint1 & -60$^{\circ}$ $\sim$ 60$^{\circ}$\\ 
                                         & Joint2 & -130$^{\circ}$ $\sim$ 130$^{\circ}$ \\ 
                                         & Joint3 & -180$^{\circ}$ $\sim$ 180$^{\circ}$ \\
                                         & Joint4 & -90$^{\circ}$ $\sim$ 90$^{\circ}$ \\
                                         & Joint5 & -90$^{\circ}$ $\sim$ 90$^{\circ}$ \\
                                         & Joint6 & -180$^{\circ}$ $\sim$ 180$^{\circ}$ \\
                                         
        \midrule 
        \multirow{4}{*}{Driving cables} & Type of cable & 1mm steel wire rope\\ 
                              & Maximum tension & 900 N \\ 
                              & Pretension force & 120 N \\ 
                              &Young's modulus & 100 Gpa\\
        \midrule 
        \multirow{3}{*}{HT-04 Motor} & Nominal torque & 13 Nm\\ 
                              & Nominal speed & 300 rpm \\ 
                              & Encoder resolution & 0.087$^{\circ}$ \\ 
        \bottomrule 
    \end{tabular}
    \label{tab:Specifications} 
\end{table}

\section{Kinematics Model Analysis}
\subsection{Forward Kinematics}

The DH coordinate system of D3-Arm is established to the standard DH method, as shown in Fig. \ref{fig:Proportion}. Due to the implementation of rolling constraints, we utilize 10 equivalent joints to compute the forward kinematics of the 6-DOF D3-Arm with four constrain equations as followed
\begin{equation}
    \begin{cases}
    \Delta \theta_2 = \Delta \theta_3, \\
    \Delta \theta_4 = \Delta \theta_5, \\
    \Delta \theta_6 = -\Delta \theta_9, \\
    \Delta \theta_7 = \Delta \theta_8
    \end{cases}
\end{equation}

Then, the forward kinematics of the 6-DOF D3-Arm can be described as
\begin{equation}
f(\theta) = \mathbf{T}_2^1 \mathbf{T}_3^2 \mathbf{T}_5^4 \cdots \mathbf{T}_{10}^{9}
\label{eq:Fk}
\end{equation}
where $\mathbf{T}_i^{i-1} (2\leq i\leq 10)$ is the local homogeneous transformation matrices between adjacent coordinate systems and can be derived from local transformation $\mathbf{R}_i^{i-1}$ and position $\mathbf{p}_i^{i-1}$
\[
\mathbf{T}_i^{i-1} = 
\begin{bmatrix}
\mathbf{R}_i^{i-1} & \mathbf{p}_i^{i-1} \\
0 & 1 
\end{bmatrix}
\]

\subsection{Inverse Kinematics}
Having 10 joints variables, 6 independent DOFs and 4 constraint equations, D3-Arm can solve the inverse kinematics by substituting an improved Jacobian matrix $\mathbf{J}_{IM}$ into the least squares method based on the pseudo-inverse of the Jacobian matrix. Based on (\ref{eq:Fk}), the Jacobian matrix of D3-Arm can be derived and written as
\begin{equation}
\mathbf{J} = 
\begin{bmatrix}
\mathbf{v}_1  & \mathbf{v}_2 & \cdots & \mathbf{v}_{10} \\
\omega_1  & \omega_2 & \cdots & \omega_{10} \\

\end{bmatrix}
\quad \in \boldsymbol{R}^{6 \times 10}
\end{equation}

According to the constraint equations, the relationship between the 10 joint angle rates $\boldsymbol{\dot{\theta}}_{10 \times 1}$ of the D3-Arm and the 6 independent joint angle rates $\boldsymbol{\dot{\theta}}_{6 \times 1}^{\prime}$ can be expressed as follows
\begin{equation}
    \boldsymbol{\dot{\theta}}_{10 \times 1} = \mathbf{U}^{\top} \boldsymbol{\dot{\theta}}_{6\times1}^{\prime}\\
\end{equation}

\begin{equation}
    \mathbf{U} = 
    \begin{bmatrix}
    1 & 0 & 0 & 0 & 0 & 0 & 0 & 0 & 0 & 0\\
    0 & 1 & 1 & 0 & 0 & 0 & 0 & 0 & 0 & 0\\
    0 & 0 & 0 & 1 & 1 & 0 & 0 & 0 & 0 & 0\\
    0 & 0 & 0 & 0 & 0 & 1 & 0 & 0 & -1 & 0\\
    0 & 0 & 0 & 0 & 0 & 0 & 1 & 1 & 0 & 0\\
    0 & 0 & 0 & 0 & 0 & 0 & 0 & 0 & 0 & 1\\    
    \end{bmatrix}
\end{equation}
where $\boldsymbol{\dot{\theta}}_{10 \times 1} = \begin{bmatrix} \dot{\theta_1} & \dot{\theta_2} & \cdots & \dot{\theta}_{10} \end{bmatrix}$ and $\boldsymbol{\dot{\theta}}_{6 \times 1}^{\prime} = \begin{bmatrix} \dot{\theta_1}&\dot{\theta_2}&\dot{\theta_4}&\dot{\theta_6} &\dot{\theta_7}&\dot{\theta}_{10}\end{bmatrix}$. Then, the inverse kinematics can be calculated as
\begin{equation}
    \mathbf{\dot{X}} = \mathbf{J}\boldsymbol{\dot{\theta}}_{10 \times 1} = \mathbf{J}\mathbf{U}^{\top}  \boldsymbol{\dot{\theta}}_{6 \times1 }^{\prime} = \mathbf{J}_{IM}\boldsymbol{\dot{\theta}}_{6 \times 1}^{\prime}
\end{equation}
\begin{equation}
    \boldsymbol{\dot{\theta}}_{6 \times 1}^{\prime} = \mathbf{J}_{IM}^{-1}\mathbf{\dot{X}} 
\end{equation}
where $\mathbf{\dot{X}}$ represent the rate of change of the end-effector.

\section{EXPERIMENT AND RESULTS}
To verify the performance of the proposed D3-Arm, a prototype is built to investigate its position repeatability, load capacity and dynamic motion. In the experiment, we used the motion capture system(V120 Trio, Optitrack) to measure the pose and velocity of the end effector of D3-Arm at 120Hz.

\subsection{Decoupling Verification}
\begin{figure}[t]
    \centering
    \includegraphics[width=1\linewidth]{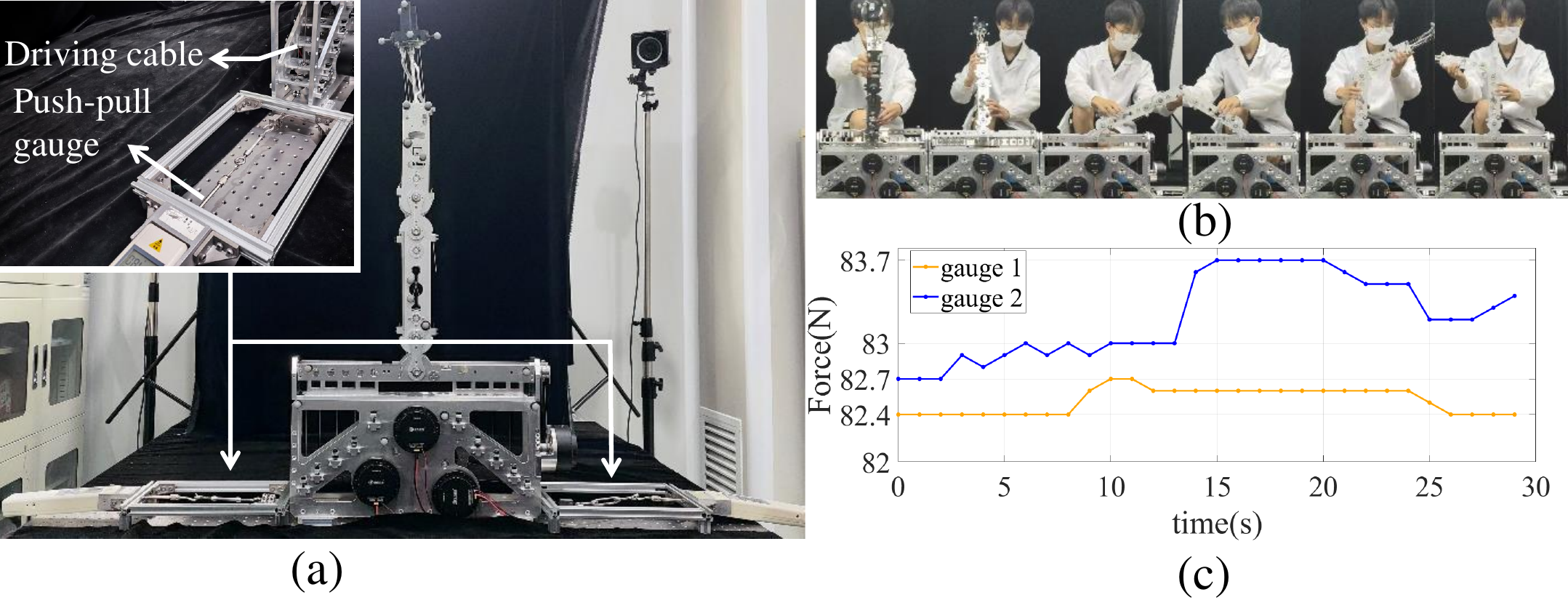}
    \caption{Decoupling verification of D3-Arm. (a) Experimental setup for force measurement. (b) Snapshots of the process of applying external force to Joint1, 2, 3. (c) The variation of cable tension measured by two Push-pull gauges. Measurement accuracy is 0.1 N.}
    \label{fig:Decoupled}
    \vspace{-0.5cm}
\end{figure}

To validate the effectiveness of the proposed decoupling mechanism, an experiment is conducted as shown in Fig.\ref{fig:Decoupled}. We remove one pair of driving cables of the end-effector from the motor's roller and connected them to a Push-pull gauge fixed to the base through the cable pre-tension mechanism, which applies a tension of 82 N. During the experiment, external forces are applied to move Joint1, Joint2, and Joint3 to simulate interference on the joints, and the variations in cable tension are recorded from the Push-pull gauge, as shown in Fig. \ref{fig:Decoupled}(b). The experimental results are shown in Fig. \ref{fig:Decoupled}(c). The maximum variation in tension on the cable is 1 N. Based on the Young's modulus illustrated from Table. \ref{tab:Specifications}, the corresponding length change does not exceed 0.01 mm, demonstrating that the motion interference of the joints integrated with decoupling mechanism can be considered negligible. This enables high-precision position control among the joints.

\subsection{Position Repeatability}
 According to the performance criteria and test methods of ISO 9283\cite{iso1998}, the repeatability test of D3-Arm is conducted with closed-loop position control at every motor. Considering that the application scenarios of this cable-driven robotic arm, whose motors and actuators are all mounted at the base, mostly do not allow for the installation of external observation sensors, the joint angle information is not included in the closed-loop control during this experiment.
 
 \begin{figure}[h]
    \centering
    \subfigure[]{
        \includegraphics[width=0.18\textwidth]{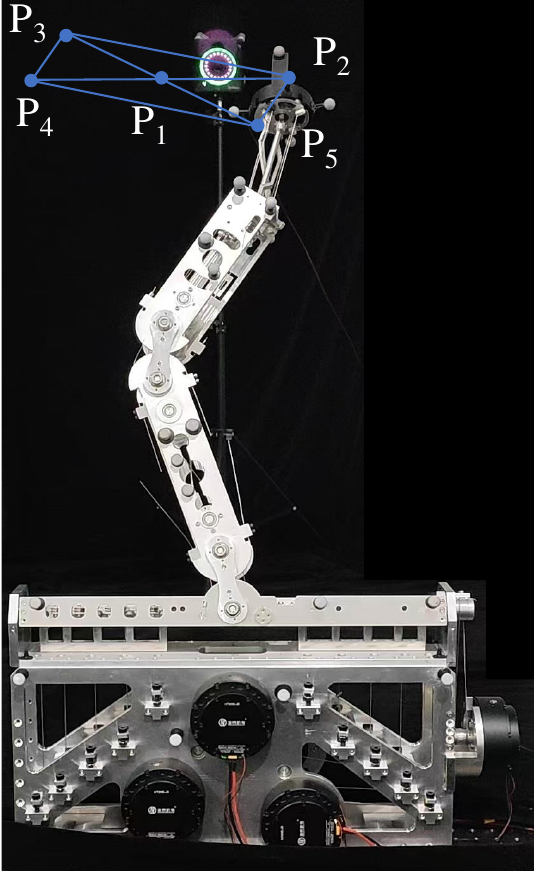}
        \label{fig:scatter1}
    }
    \subfigure[]{
        \includegraphics[width=0.26\textwidth]{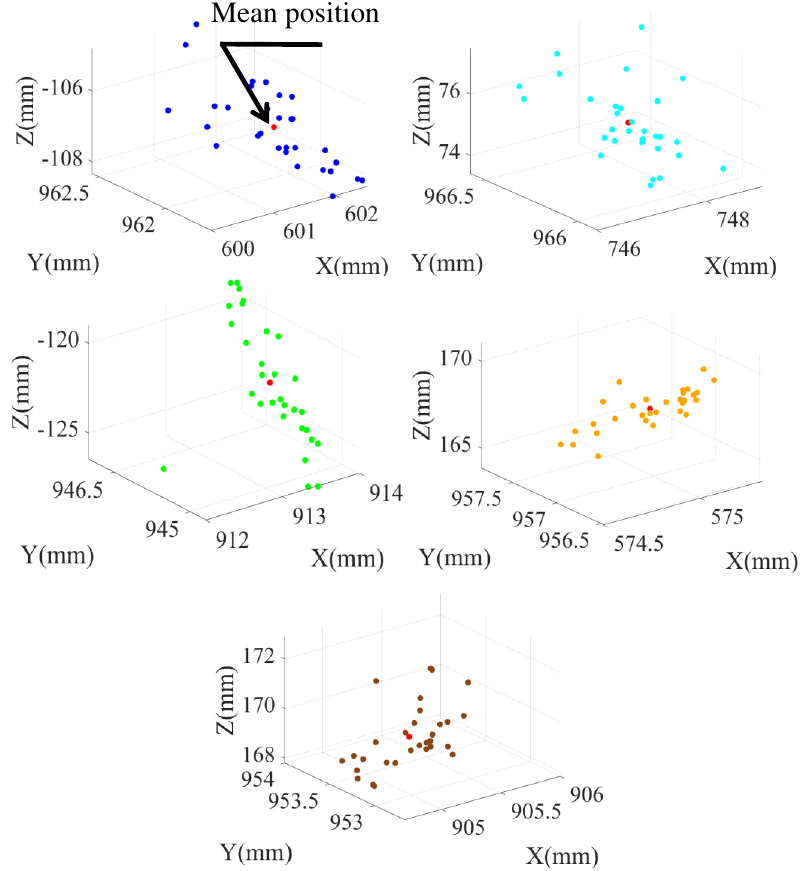}
        \label{fig:scatter2}
    }
    \caption{Repeatability test based on ISO 9283:1998. (a) 5 pre-defined target positions(P$_1$,P$_2$,P$_3$,P$_4$,P$_5$) for the test. D3-Arm successively moves between these 5 positions and repeats for 30 times. (b) Repeatability test results at 5 pre-defined target positions. Measurement accuracy is $\pm$ 0.2 mm.}
    \label{fig:repeatability}
    \vspace{-0.2cm}
\end{figure}

 The position repeatability test of D3-Arm's end-effector is measured at 5 pre-defined positions as shown in Fig. \ref{fig:scatter1}. D3-Arm successively moves to the poses P$_1$, P$_2$, P$_3$, P$_4$, P$_5$ and repeats for 30 times. The measurement point cloud results at these points are shown in Fig. \ref{fig:scatter2} and the average distance (Mean), its standard deviation (STD.DEV) and 3-sigma distance between the mean and recorded positions were summarized in Table \ref{tab:rp}. The average position error of D3-Arm is 1.2896mm, which is lower than the 4.9mm of Twist Snake\cite{snake} and other cable-driven arms with all the motors located at the base\cite{low-cost}, proving the effectiveness of decoupling mechanisms in improving control accuracy performance. 

\begin{table}[h]
    \centering
    \caption{Positioning repeatability test results}
    \begin{tabular}{cccc} 
        \toprule 
        Pose & Mean(mm) & STD.DEV.(mm) & 3-SIGMA(mm) \\ 
        \midrule 
        $P_1$ & 0.8562 & 0.4902 & 2.3269 \\ 
        $P_2$ & 1.9868 & 1.1983 & 5.5817 \\ 
        $P_3$ & 1.0997 & 0.7271 & 3.2810 \\ 
        $P_4$ & 1.5977 & 1.0693 & 4.8056 \\ 
        $P_5$ & 0.9078 & 0.6749 & 2.9324 \\ 
        \midrule 
        Total & 1.2896 & 0.8320 & 3.7855 \\ 
        \bottomrule 
    \end{tabular}
    \label{tab:rp}
\end{table}

 The 3-sigma distance, which corresponds to the positioning repeatability of ISO 9283, is 3.7855 mm. Due to the results of decoupling verification, we analyze that the error mainly comes from the lack of the joint sensors and the elasticity of cables, which reaches 1041mm length for the end-effector. 

\subsection{Load Capacity}
The load capacity is one of the important performance indicators of a robotic arm; however, it is noteworthy that the design objective of this cable-driven arm does not prioritize high load capacity. The load test is mainly conducted on Joint1 since the the load on the entire arm is mainly concentrated on Joint1.

\begin{figure}[t]
    \centering
    \includegraphics[width=1\linewidth]{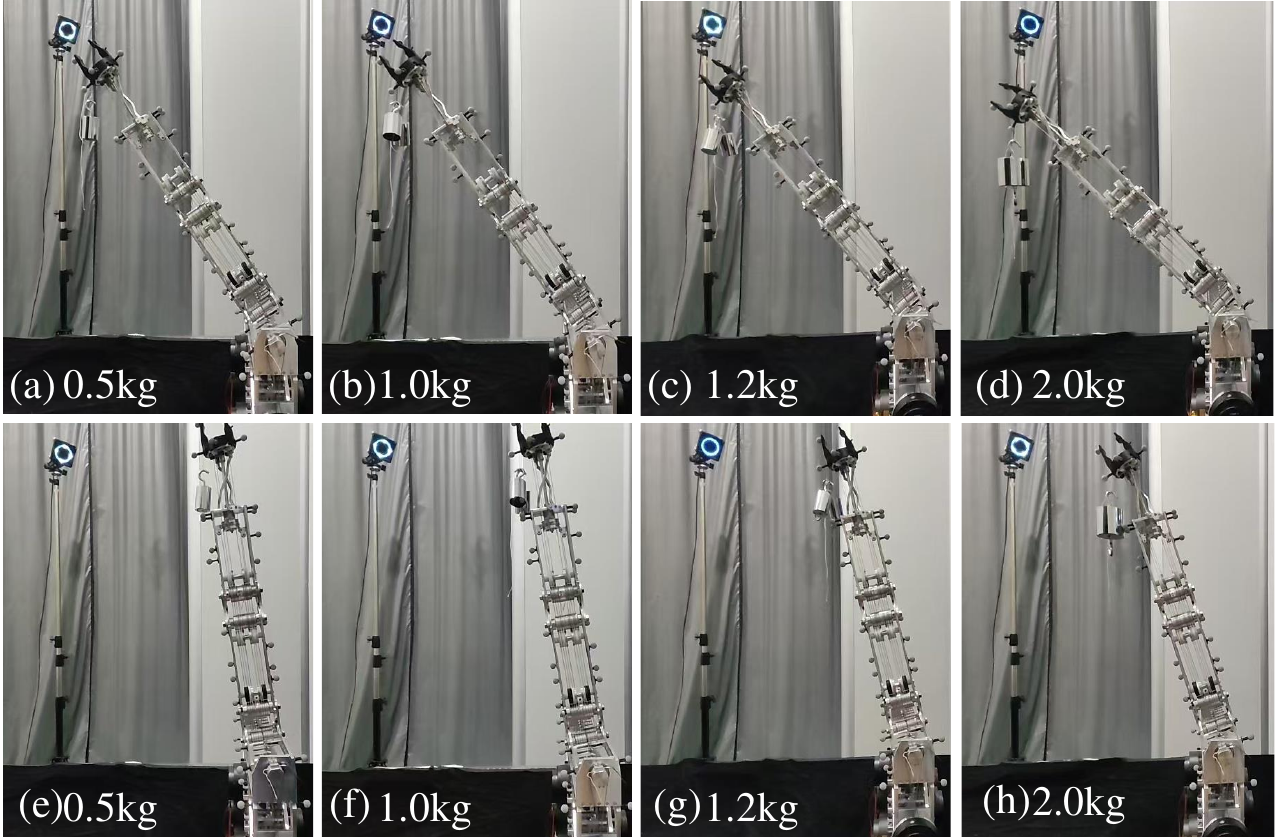}
    \caption{Load capacity experiment of D3-Arm. (a-d) The payload tests where Joint1 bears loads of 0.5-2.0 kg at angle of -50$^{\circ}$. (e-h) The payload tests where Joint1 bears loads of 0.5-2.0 kg at angle of 0$^{\circ}$.}
    \label{fig:load}
    \vspace{-0.3cm}
\end{figure}

During the testing process, we hang weights ranging from 1.0 kg to 2.0 kg on the end effector and drove the Joint1 to move from -50$^{\circ}$ to 0$^{\circ}$. The sequential images of the test is presented in Fig. \ref{fig:load}. From the test, it can be seen that D3-Arm can smoothly carry a weight of up to 2.0 kg. The cables began to detached from the cable-termination-mechanism under a heavier weight, which limits the load-bearing capacity of D3-Arm. Notably, as the weight increases, the end-effector of D3-Arm begins to exhibit significant deformation as shown in Fig. \ref{fig:load}(a-h), demonstrating the low stiffness of the robotic arm. This drawback is attributed to the elasticity of the long-distance and thin cables, which can be improved by incorporating a variable stiffness mechanism. 
\subsection{Dynamic Motion}
\begin{figure}[t]
    \centering
    \includegraphics[width=1\linewidth]{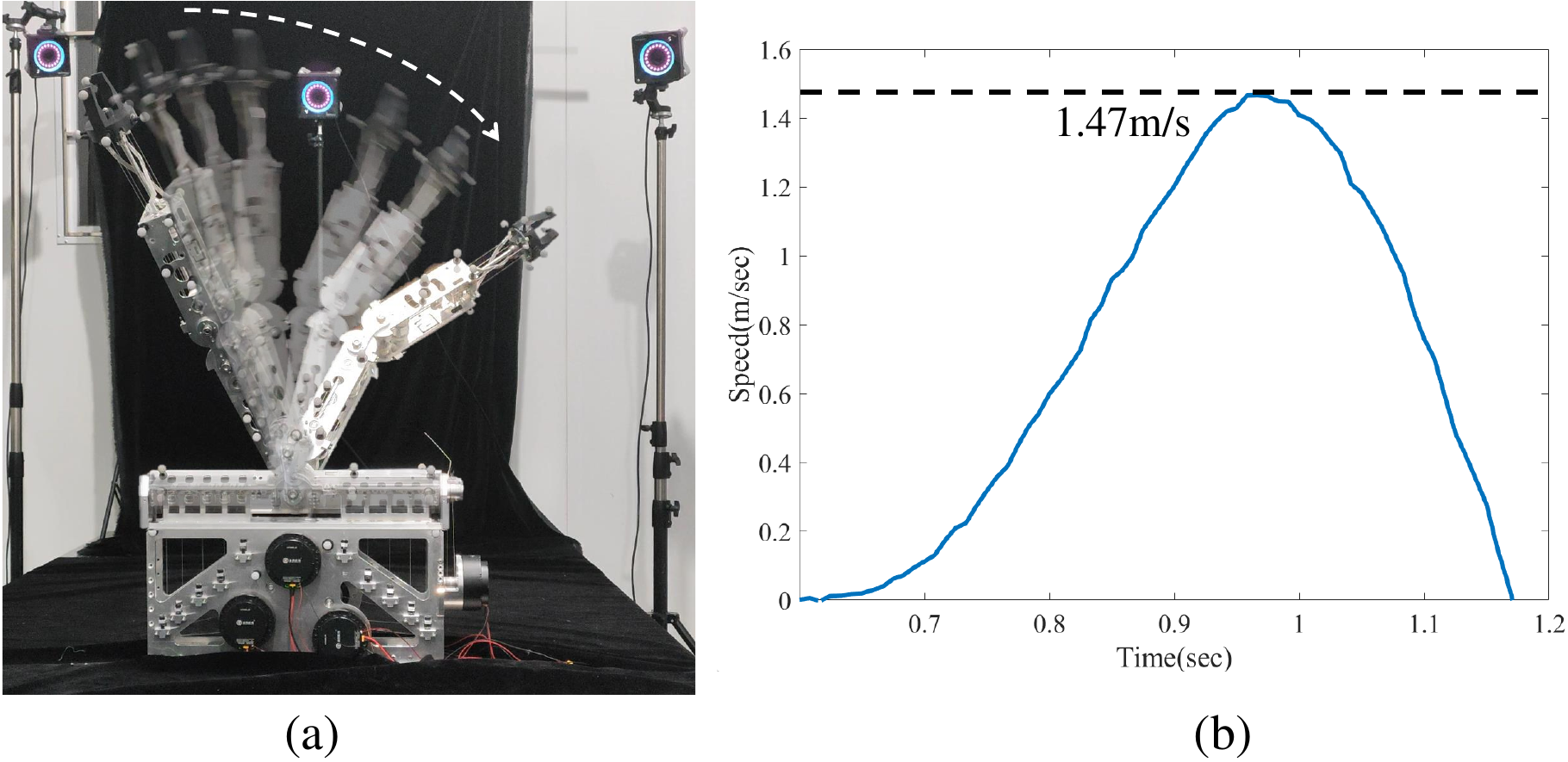}
    \caption{High speed motion test. (a) Snapshots of the test motion. The white dashed arrow indicates the direction of D3-Arm's movement. (b) Speed graphs of the motion.}
    \label{fig:high-dynamic}
    \vspace{-0.5cm}
\end{figure}

To evaluate the high-speed motion capabilities of the D3-Arm, we design a high-speed swinging experiment using the whole arm. Each joint of the D3-Arm is controlled using PID position control and moves along the predefined trajectories within the joint space. The sequential images and end-effector velocity, which is measured by the motion capture system, are illustrated in Fig. \ref{fig:high-dynamic}. In the dynamic motion experiment, the D3-Arm executed rapid swings within its workspace, reaching a maximum movement speed of 1.47 m/s and an acceleration of up to 10.55 m/s$^2$.

\section{CONCLUSIONS}
In this paper, we develop a fully-decoupled and lightweight cable-driven robotic arm system named D3-Arm with all its electrical components such as motors placed at the base. The cables are aligned and transmitted efficiently through decoupling mechanisms composed by grooved bearings and pulleys, significantly enhancing the control precision compared to other cable-driven robotic arms with all the motors positioned at the base. By integrating the remote driving feature of cables and fully-decoupled transmission mechanisms, the D3-Arm is capable of performing tasks in various environments including underwater and high-radiation that require isolation and protection of electrical components. In future work, we plan to incorporate the variations of cable tension into the control of the entire arm to further improve the control accuracy and performance of D3-Arm.

\addtolength{\textheight}{-5cm}   



\newpage
\bibliographystyle{ieeetr}
\bibliography{main}

\end{document}